\documentclass[conference]{IEEEtran}
\IEEEoverridecommandlockouts
\usepackage{cite}
\usepackage{amsmath,amssymb,amsfonts}
\usepackage{algorithmic}
\usepackage{graphicx}
\usepackage{textcomp}
\usepackage{xcolor}
\usepackage{bbm}
\usepackage{multicol}
\usepackage{multirow}
\usepackage{makecell}
\usepackage{subfigure}
\usepackage{soul}
\usepackage{enumitem}

\usepackage{bigdelim}
\usepackage{hyperref}

\usepackage{booktabs}
\usepackage{pifont}%

\newcommand{\xmark}{\ding{55}}
\def\BibTeX{{\rm B\kern-.05em{\sc i\kern-.025em b}\kern-.08em
    T\kern-.1667em\lower.7ex\hbox{E}\kern-.125emX}}

\makeatletter
\newcommand{\linebreakand}{%
  \end{@IEEEauthorhalign}
  \hfill\mbox{}\par
  \mbox{}\hfill\begin{@IEEEauthorhalign}
}

\makeatother

\begin{document}

\title{A Scalable and Transferable Time Series Prediction Framework for Demand Forecasting}

\author{\IEEEauthorblockN{Young-Jin Park\IEEEauthorrefmark{2}}
\IEEEcompsocitemizethanks{\IEEEcompsocthanksitem\IEEEauthorrefmark{2}This work has been completed while the authors were working at NAVER.}
\IEEEauthorblockA{
\textit{MIT LIDS}\\
Cambridge, MA, USA \\
youngp@mit.edu
}
\and
\IEEEauthorblockN{Donghyun Kim\IEEEauthorrefmark{1},\IEEEauthorrefmark{2}}
\IEEEcompsocitemizethanks{\IEEEcompsocthanksitem\IEEEauthorrefmark{1}contributed equally and share joint second authorship.}
\IEEEauthorblockA{
\textit{Seoul National University}\\
Seoul, South Korea \\
leok0909@snu.ac.kr}
\and
\IEEEauthorblockN{Fr{\'e}d{\'e}ric Odermatt\IEEEauthorrefmark{1},\IEEEauthorrefmark{2}}
\IEEEauthorblockA{
\textit{ETH Z{\"u}rich}\\
Zürich, Switzerland \\
odermafr@ethz.ch}
\linebreakand
\IEEEauthorblockN{Juho Lee\IEEEauthorrefmark{2}}
\IEEEauthorblockA{
\textit{Superpetual Inc.}\\
Seoul, South Korea \\
3juho@superpetual.com}
\and
\IEEEauthorblockN{Kyung-Min Kim}
\IEEEauthorblockA{
\textit{NAVER CLOVA}\\
\& \textit{NAVER AI Lab}\\
Seongnam, South Korea \\
kyungmin.kim.ml@navercorp.com}
}

\maketitle

\begin{abstract}
Time series forecasting is one of the most essential and ubiquitous tasks in many business problems, including demand forecasting and logistics optimization.
Traditional time series forecasting methods, however, have resulted in small models with limited expressive power because they have difficulty in scaling their model size up while maintaining high accuracy.
In this paper, we propose \textbf{F}orecasting \textbf{orchestra} (\textbf{Forchestra}), a simple but powerful framework capable of accurately predicting future demand for a diverse range of items.
We empirically demonstrate that the model size is scalable to up to 0.8 billion parameters.
The proposed method not only outperforms existing forecasting models with a significant margin, but it could generalize well to unseen data points when evaluated in a \emph{zero-shot fashion} on downstream datasets.
Last but not least, we present extensive qualitative and quantitative studies to analyze how the proposed model outperforms baseline models and differs from conventional approaches.
The original paper was presented as a full paper at ICDM 2022 and is available at: \href{https://ieeexplore.ieee.org/document/10027662}{https://ieeexplore.ieee.org/document/10027662}.
\end{abstract}

\section{Introduction}
\begin{figure*}[ht]
  \centering
  \includegraphics[width=\linewidth]{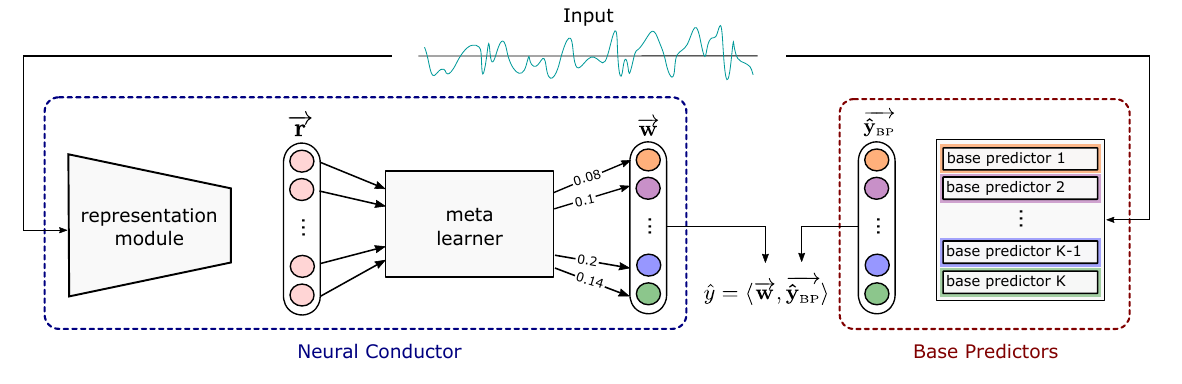}
  \caption{The network architecture of Forchestra. Forchestra consists of $K$ base predictors and a neural conductor. For a given time series, each base predictor outputs its forecast and the meta learner scores the importance weight of each base predictor based on the representation vector inferred by the representation module. All modules in Forchestra are jointly trained in an end-to-end manner.}
  \label{fig:teaser}
\end{figure*}

Demand forecasting is a crucial component of supply chain management; to make better inventory planning decisions and maximize revenue, both E-commerce and offline retail stores require accurate forecasting of future demand for products they offer \cite{fildes2019retail}.
As such, several sales forecasting competitions, such as the M competitions \cite{makridakis1982accuracy, makridakis1993m2, makridakis2000m3, makridakis2018m4, MAKRIDAKIS2021}, have been held over the past decades, and various forecasting methods have been presented for more accurate and robust time series prediction.

Deep learning-based algorithms have recently gained popularity due to their flexibility, which allows them to be applied to a wide range of time series without requiring handcrafted features or additional training on new products \cite{MAKRIDAKIS2021}.
However, their performance is often even worse than classical algorithms due to their tendency to overfit \cite{brownlee2019comparing, jung2020worrying}.
Likewise, those methods struggle to scale up their model size effectively, and in this paper we also empirically observe that a single large model with hundreds of layers or thousands of hidden states performs poorly when forecasting on our datasets.

Very recently, several studies have attempted to build \emph{foundation models \cite{bommasani2021opportunities} for time-series forecasting \cite{das2023decoder, garza2023timegpt, darlow2023dam}}.
These foundation models have tried to build a large-scale forecasting model to generalize to unseen data in a zero-shot manner rather than building instance-specific forecasting.
As such, authors believe this work resonates with a such burgeoning trend, establishing itself as one of the pioneering endeavors to conceive a scalable and transferable framework (for demand forecasting).

Meanwhile, ensemble approaches combining classical and deep learning methods have seen a relatively high amount of success in the time series forecasting domain, partly attributed to their increased robustness to the distribution shift; since time series prediction is a task that requires inference on future where no data has been observed, a distribution shift between the training, validation, and test periods is inevitable \cite{oliveira2015ensembles}.

Despite the benefits of ensemble learning, selecting and assigning base models from the pool remains a challenging task \cite{gastinger2021study}.
If ensembling methods are not applied carefully, they may run the risk of smoothing or disharmonizing the forecasting results, and lowering forecast accuracy.
In addition, for problems like demand forecasting, where the number of time series to be forecasted is in the order of hundreds of thousands, the issues become even more pronounced because each time series instance has different temporal characteristics that a model needs to capture and extend into the future.

Moreover, conventional ensemble methods often rely on a user-defined set of base models individually that have been trained.
According to \cite{fort2019deep}, however, an ensemble using deep learning models is able to improve its accuracy and out-of-distribution robustness when each base model is trained to explore different modes in the function space.
As such, this paper is shaped by investigating the question: How can we construct a large number of distinct but jointly-working base models and guide them in a more flexible manner?
To address this question, we present a scalable and flexible time series prediction framework, illustrated in Figure~\ref{fig:teaser}.

The rest of the paper is organized as follows:
In Section~\ref{sec:method}, we describe how the proposed framework works and is trained in detail.
Section~\ref{sec:related} introduces literature in demand forecasting, ensemble forecasting, and representation learning for time series.
Finally, in Section~\ref{sec:results} and Section~\ref{sec:discussion}, we present extensive qualitative and quantitative studies to analyze the proposed model in depth, using two large-scale real-world demand forecasting datasets, \texttt{E-commerce} and \texttt{M5}.

\section{Methodology} \label{sec:method}

\subsection{Problem Definition}

Given a training dataset of $N$ time series of length $T$: $\mathcal{D}_{\mathrm{train}} = \{ \mathbf{x}^{(i)}_{1:T}\}_{i=1}^{N}$ where $\mathbf{x}^{(i)}_t = [y^{(i)}_t, \mathbf{u}^{(i)}_t] \in \mathbb{R}^{1+F}$.
$y^{(i)}_t \in \mathbb{N}_0$ and $\mathbf{u}^{(i)}_t \in \mathbb{R}^F$ are a sequence of past observations (e.g., daily sales) and a sequence of features (e.g., availability status), respectively, of the $i$th instance at time $t$.
The aim of demand forecasting is to predict the time series for the next $P$ time steps: $\{y^{(i)}_{T+1:T+P}\}_{i=1}^{N}$.
For the sake of simplicity, we drop the superscript $i$ for the following sections.

\subsection{Forchestra}

This paper proposes a simple but powerful forecasting framework, \textbf{\textit{Forchestra}}.
Forchestra consists of two parts: 1) base predictors (Section~\ref{sec:bp}) and 2) a neural conductor (Section~\ref{sec:nc}).
For a given time series, each \emph{base predictor} outputs its respective forecast based on historical observations.
On top of the base predictors, the \textit{neural conductor} adaptively assigns the importance weight for each predictor by looking at the representation vector provided by a representation module.
Finally, Forchestra aggregates the predictions by the weights and constructs a final prediction.
In contrast to previous forecasting approaches, the neural conductor and all base predictors of Forchestra are trained in an end-to-end manner; this allows each base predictor to modify its reaction to different inputs, while supporting other predictors and constructing a final prediction jointly.

\subsubsection{Base Predictors (BPs)}
\label{sec:bp}

This section introduces the first principal component in the proposed method, a set of K base predictors $\mathbf{f}_\theta$. Each base predictor is represented as a mapping $f_{\theta_k}: \mathbb{R}^{C \times (F+1)} \mapsto \mathbb{R}^{P}$ parameterized by trainable weights $\theta_k$, and maps historical data over a context window $C$ to a prediction window of length $P$. More precisely, the prediction of a single base predictor can be described as:
\begin{equation}
    \hat{y}_{k, t+1:t+P} = f_{\theta_k}(\mathbf{x}_{t-C-1:t}) \in \mathbb{R}^P .
\end{equation}
However to increase readability we refer to the joint output of the base predictors as:
\begin{equation}
    \mathbf{\hat{y}}_{BP} = \begin{bmatrix}
\hat{y}_{1,t+1:t+P}\\
\vdots\\
\hat{y}_{K,t+1:t+P}
\end{bmatrix} \in \mathbb{R}^{K \times P} . 
\end{equation}
Unlike methods comprised of a single model, such as Prophet \cite{taylor2018forecasting} and DeepAR \cite{salinas2020deepar}, %
our proposed framework consists of $K$ distinct predictors, with $\theta_k$ as the learnable parameters.

The authors have pondered the qualities that a good set of BPs should have in order to function effectively.
Of course, as in general ensemble approaches, any BP by itself must possess decent predictive ability to some degree in order to be useful to the ensemble. 
However, even if all BPs have excellent predictive power, the margin of improvement when they are combined might be small if they have similar predictive behavior. 
In light of this, we hypothesized that, even if each BP's performance is low, the synergy between the BPs can be further strengthened when BPs effectively operate together. 
Discussion on the BP's role is analyzed in Section~\ref{sec:discussion}.

\subsubsection{Neural Conductor (NC)} \label{sec:nc}

To make the most of the capability of competent base predictors, it is also essential to have a high-level conductor that can guide those predictors and provide a unified view of the forecasts.
With this in mind, we introduce the next key component, a neural conductor (NC).

The first and most important function of the neural conductor is to produce an expressive representation of a given time series such that the neural conductor is able to activate the base predictors flexibly.
As such, this paper introduces a neural network $g_{\phi_r}: \mathbb{R}^{W \times (F + 1)} \mapsto \mathbb{R}^D$ parameterized by $\phi_r$ that maps historical data to a latent representation for each time series, extracting a representation vector:
\begin{equation}
    \mathbf{r_t} = g_{\phi_r}(\mathbf{x}_{t-W-1:t}) \in \mathbb{R}^D
\end{equation}
where $W$ is a window size. We refer to $g_{\phi_r}$ as \emph{representation module} from here on. 
Note that we refer to $\mathbf{r}_t$ by $\mathbf{r}$ to increase readability and that the window size $W$ must not necessarily be equal to the context length $C$.
The representation is expected to capture important characteristics that can identify a given time series and distinguish it from other instances or time points.

Based on its representation, a trainable meta learner $h_{\phi_w}: \mathbb{R}^D \mapsto \mathbb{R}^K$ parameterized by $\phi_w$ extracts the importance weights for each base predictor:
\begin{equation} \label{eq:meta}
    \mathbf{w} = \{w_1, \cdots, w_K\} =  \text{Softmax} \left( h_{\phi_w}(\mathbf{r}) \right) \in \mathbb{R}^K .
\end{equation}

The final prediction of the Forchestra is obtained by aggregating the predictions from each base predictor based on the weights computed by the neural conductor:
\begin{equation}
    \hat{y} = \sum_{k=1}^{K} w_k \mathbf{\hat{y}}_{\text{BP}_k} \in \mathbb{R}^P .
\end{equation}

Finaly, the network parameters, $\{\phi_r, \phi_w, \theta_1, \ldots, \theta_K \}$, are learned to minimize the L1-norm (i.e., MAE) between the final prediction and ground truth:
\begin{equation}
    \mathcal{L} = \sum_i \sum_t \lVert y^{(i)}_{t+1:t+P} - \hat{y}^{(i)}_{t+1:t+P} \rVert_1
\end{equation}

In this paper, we used LSTMs for base predictors, a dilated CNN for the representation module, and a fully-connected layer for the meta learner.
See Appendix~\ref{sec:hyper} for more details.

\subsection{Pre-Training a Neural Conductor}

From the viewpoint of deep generative modeling, the neural conductor acts as the encoder, and $\phi_r$ and $\phi_w$ are optimized to output compact representations that can distinguish time series with dissimilar patterns.
Following the observation, we introduce a self-supervised pre-training strategy for a neural conductor.
Despite the fact that any method could be applied, the TS2Vec \cite{yue2021ts2vec}, one of the state-of-the-art self-supervised contrastive learning methods, is applied in this paper.

First, time series are augmented by both cropping and masking. 
Instead of feeding a single value of a time series at a timestamp values are projected to a higher dimensional space using an embedding layer. 
As every timestep gets its own representation, the embedding over a period is aggregated by applying max-pooling over the single timestep representations in the time dimension. 
For a time series of $N$ time units there are $\lfloor\log_2 N\rfloor$ levels of pooling in this approach. 
In every batch the masking and cropping augmentations are applied separately twice to every sample, producing two partially overlapping and differently masked time series snippets. 
Let $r$ and $r'$ be these two masked and cropped representations and $\Omega$ the set of time indices in a period we are learning on, on a hierarchical level we are currently working on.

\begin{equation}
  \resizebox{0.91\columnwidth}{!}{%
      $\mathcal{L}_{\text{temporal}}^{(i,t)} = - \log \Bigg[\frac{\exp(r_{i,t} \cdot r'_{i,t})}{\sum_{t' \in \Omega} \big(\exp(r_{i,t}\cdot r'_{i,t'}) + \mathbbm{1}_{[t\neq t']} \exp(r_{i,t} \cdot r_{i,t'})\big)}\Bigg]$%
  }
\end{equation}

For the \textit{temporal} loss the model should maximize the inner product between the representations of the same instance (sample) at the same timestamp (just cropped and masked differently) while minimizing the similarity to the representations at other timesteps of both augmentations.

\begin{equation}
  \resizebox{0.90\columnwidth}{!}{%
      $\mathcal{L}_{\text{instance}}^{(i,t)} = - \log \Bigg[\frac{\exp(r_{i,t} \cdot r'_{i,t})}{\sum_{j=1}^{B} \big(\exp(r_{i,t}\cdot r'_{j,t}) + \mathbbm{1}_{[i\neq j]} \exp(r_{i,t} \cdot r_{j,t})\big)}\Bigg]$%
      }
\end{equation}

For the \textit{instance} loss the model should maximize the inner product between the representations of the same instance at the same timestamp while minimizing the similarity to all other instances in the same batch.

For a batchsize of $B$ and $P$ \textit{stages} of pooling let $t_p, \: p \in \{1,..,P\}$ denote the amount divisions of the time axis at every pooling level.
Through the iterative max pooling $t_p$ becomes smaller with every pooling step, following an exponential decay.
Finally, the neural conductor is pre-trained by minimizing the following the hierarchical loss function:
\begin{equation}
    \mathcal{L}_{pre} = \frac{1}{2 P B} \sum\limits_{\substack{p \in \{1,...,P\} \\ i \in \{1,...,B\} \\ t \in \{1,...,t_p\}}} \frac{1}{t_p} \Big( \mathcal{L}_{temporal}^{(i,t)} + \mathcal{L}_{instance}^{(i,t)} \Big) ~~ 
\end{equation}

\section{Related Work} \label{sec:related}

\subsection{Time Series Forecasting Overview}
The goal of forecasting time series data, which is characterized by trend, seasonality, stochasticity, heteroscedasticity, etc., is to predict how sequential data extends into the future. 

It is natural to assume that bigger and more complex models will provide better forecasting power. 
However, the "M" competitions \cite{makridakis1982accuracy, makridakis1993m2, makridakis2000m3, makridakis2018m4, MAKRIDAKIS2021}, one of the most famous competitions focused on time series forecasting, have shown that this is not always true. 
It was not until the fourth iteration of M competitions (M4) that sophisticated machine learning (ML) based models started to beat traditional approaches. 
However, only two ML based methods were more accurate than the classical statistics-based models while the majority of spots at the top of the ranking were still taken by classical models.

Finally, in the M5 competition \cite{MAKRIDAKIS2021}, where the target values are retail sales from Walmart, ML models started to assert their dominance.
The top 50 ranking models were all ML based ones and their average performance beat the most accurate statistical model by 14\%.
Therefore the M5 competition is the first competition where top-performing models are both pure ML and significantly better than the statistical models.
Promising results from state-of-the-art deep learning implementations such as DeepAR \cite{salinas2020deepar} and N-BEATS \cite{oreshkin2019n} have further motivated research in this direction.

\subsection{Ensemble Forecasting}
Ensemble methods have found a significant degree of success in the domain of time series forecasting and demonstrated superior performance in forecasting competitions \cite{timmermann2006forecast, makridakis2018m4, montero2020fforma}.
The flexible selection of an appropriate set of weights for the underlying base models is one of the most critical challenges that ensemble approaches have attempted to resolve \cite{gastinger2021study}.
Conventionally, ensemble methods simply average base model predictions for the final forecast or adjust weights based on the validation score of base models.
\cite{cerqueira2017dynamic} select the top $K$ base models and weigh them based on the performance of the most recent data points.
\cite{pawlikowski2020weighted} tune weights using heuristic functions (e.g., an inverse of the performance score) and rank 3rd in the M4 competition.
\cite{montero2020fforma}, the runner-up of the M4 competition, use an ensemble method that weighs base models in the pool by using handcrafted time series features \cite{hyndman2019tsfeatures}.
Additionally, there have been some studies that have focused on the selection of base models; in forward \cite{caruana2004ensemble} or eliminating models in backward \cite{pawlikowski2020weighted}, based on their validation error.

Despite these efforts, most existing ensemble approaches are built upon a pool of user-defined models and do not further optimize the base models in order to produce the best ensemble result. 
Additionally, these methods choose the model that is most likely to perform the best in the future, rather than effectively combining models.
However, as discussed in Section~\ref{sec:discussion}, such a strategy does not guarantee the best.

The most similar approach to the proposed method would be the mixture-of-experts (MoE) \cite{zeevi1996time}.
However, the previous work has not clearly introduce the way to train a more effective and scalable gating network (i.e., a neural condcutor).
As illustrated in Section~\ref{sec:effectinit}, the performance of the final prediction could be improved with the proposed self-supervised pre-training, which is one of the contributions of this paper.

\subsection{Unsupervised Representation Learning for Time Series}
Following the success of unsupervised learning in computer vision \cite{chen2020simple, he2020momentum} and natural language processing \cite{gao2021simcse}, different unsupervised learning frameworks for time series data have been proposed that try to make efficient use of large amounts of unlabeled time series data and produce powerful representations \cite{franceschi2019unsupervised, ijcai2021-324, yue2021ts2vec}.

In one of the earliest applications of unsupervised learning \cite{franceschi2019unsupervised} propose a scalable representation learning framework for time series by applying a triplet loss to positive samples from a timeseries' subseries and negative samples from other instances. 
The encoder employs exponentially dilated causal convolutions to capture information over inputs of arbitrary length, setting a trend for the majority of encoders in recent works on unsupervised representation learning for time series.

\cite{yue2021ts2vec} extend this approach with the following ideas: they remove the causality part in the dilated convolution, calculate the loss at different levels in the hierarchy and use the contrastive loss \cite{hadsell2006dimensionality} instead of the triplet loss, whereby augmented context views can be contrasted both across instances and across the time dimension.
Time series are augmented by both cropping and masking.
\cite{ijcai2021-324} propose an alternative approach whereby strong and weak augmentations are fed through an encoder to then predict each other's latent space representations in a cross-view prediction task in a contrastive learning fashion while also minimizing an instance-wise contrastive loss.

Notably, all these methods can provide a scalable, expressive, and transferable (i.e., universal) representation, and they performed particularly well in time series classification problems.

\section{Experiments} \label{sec:results}

\begin{figure}[t]
    \centering
    \subfigure[\texttt{E-Commerce}]
    {
        \includegraphics[width=0.45\columnwidth]{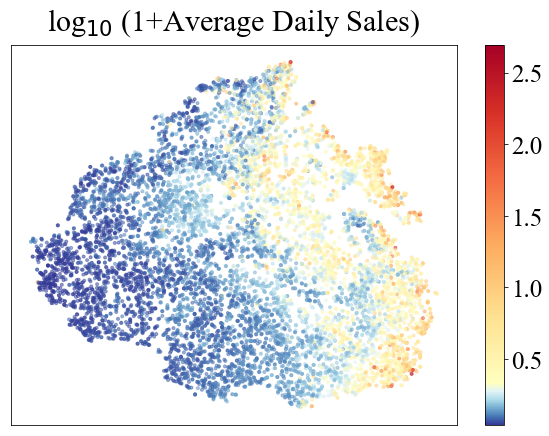}
    }
    \subfigure[\texttt{M5}]
    {
        \includegraphics[width=0.45\columnwidth]{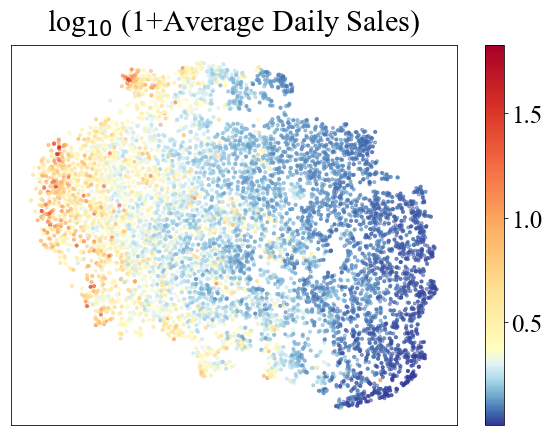}
    }
    \subfigure[\texttt{E-Commerce}]
    {
        \includegraphics[width=0.45\columnwidth]{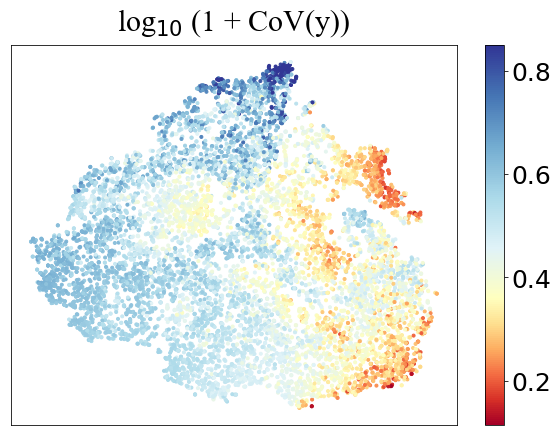}
    }
    \subfigure[\texttt{M5}]
    {
        \includegraphics[width=0.45\columnwidth]{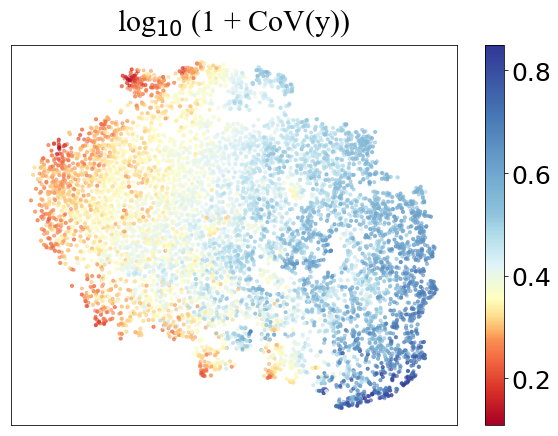}
    }
    \caption{t-SNE visualization of the representations of Forchestra for time series from \texttt{E-Commerce} and \texttt{M5} datasets, colored by the value of the title.
    }
  \label{fig:tsne}
\end{figure}

\subsection{Datasets}
We evaluate our model and baselines on our proprietary \texttt{E-Commerce} and \texttt{M5}\footnote{\url{https://www.kaggle.com/c/m5-forecasting-accuracy/data}}\cite{MAKRIDAKIS2021} datasets.
Both datasets are large-scale non-zero daily time series datasets, with \texttt{E-Commerce} and \texttt{M5} containing sales histories of $105,336$ products over $1,035$ days and $30,490$ products over $1,941$ days, respectively.
To assess the transferability in Section~\ref{sec:transfer}, we randomly selected $5 \%$ of products ($5,310$ and $1,524$, respectively) from each dataset as a transfer task and did not train on them.
For simplicity's sake, the experiment used only sales history and availability status data (i.e., whether the product was on sale or not), which are the two most critical time series features in demand forecasting.

Following the competition, the prediction length of \texttt{M5} is set to $28$ days.
For the \texttt{E-Commerce} dataset, we use a prediction length of $7$ days to align with our business logic.
However, because a single test period of \texttt{E-Commerce} is insufficiently long to adequately evaluate the performance \cite{jung2020worrying}, we backtest on four consecutive test periods (i.e., $28 = 7 \times 4$ days in total).
In short, for both datasets, we kept the last $28$ days as test period and the preceding $28$ days as validation period to select the hyperparameters.

\begin{table*}[t!]
\centering

\caption{Comparison of forecast accuracy between baselines and ours. The highest scores are highlighted in bold. The standard deviation of each metric is reported within round brackets. ARIMA and Prophet are not reported since they were not scalable.}
\label{tab:comp_train}
{\small
\begin{tabular}{ll|ccc|ccc}
\toprule
\multirow{2}{*}{}         & \multirow{2}{*}{Model} & \multicolumn{3}{c|}{\texttt{E-Commerce}} & \multicolumn{3}{c}{\texttt{M5}}  \\
                          &                        & MASE & MAE & RMSE         & MASE & MAE & RMSE       \\
\midrule
\multirow{3}{*}{Local} & ARIMA                  & \xmark & \xmark & \xmark & \xmark & \xmark & \xmark \\
                            & Prophet                & \xmark & \xmark & \xmark & \xmark & \xmark & \xmark \\
                            & SMA & 1.140 (2.95) & 1.260 (6.74)  & 1.477 (6.45) & 1.084 (1.60) & 1.099 (2.05)  & 1.420 (1.84) \\
\midrule
\multirow{6}{*}{Global} & DeepAR & 0.927 (2.99) & 1.004 (6.06)  & 1.331 (5.79) & 0.943 (1.89) & 1.056 (2.27)  & 1.550 (1.97) \\
                             & MQCNN & 0.908 (2.91) & 0.965 (5.91)  & 1.267 (5.65) & 0.907 (1.77) & 0.977 (2.01)  & 1.403 (1.74) \\
                             & MLP & 0.908 (2.90) & 0.984 (5.83)  & 1.304 (5.57) & 0.906 (1.79) & 0.987 (2.08)  & 1.456 (1.79) \\
                             & LSTM & 0.920 (2.89) & 1.002 (5.97)  & 1.300 (5.71) & 0.894 (1.82) & 0.968 (2.05)  & 1.447 (1.75) \\
                             & Transformer & 1.030 (2.98) & 1.187 (6.29)  & 1.494 (6.00) & 0.909 (1.85) & 0.999 (2.14)  & 1.482 (1.84) \\
                             & TS2Vec  & 1.108 (2.77) & 1.156 (5.84)  & 1.364 (5.60) & 1.074 (1.55) & 1.071 (1.91)  & \bf 1.378 (1.70) \\
\midrule
\multirow{3}{*}{Ensemble} & Ensemble (Best)   & 0.900 (2.89) & 0.945 (5.68)  & 1.251 (5.43) & 0.893 (1.80) & 0.968 (2.05)  & 1.439 (1.75) \\
                          & Deep Ensembles   & 0.944 (2.85) & 0.977 (5.87)  & 1.262 (5.61) & 0.893 (1.78) & 0.963 (2.00)  & 1.417 (1.71) \\
                          & N-BEATS & 0.888 (2.89) & 0.926 (5.60)  & 1.242 (5.36) & 0.890 (1.79) & 0.959 (2.00)  & 1.419 (1.70) \\
\midrule
\bf Ours                      & \bf Forchestra              & \bf 0.880 (2.87) & \bf 0.920 (5.69)  & \bf 1.233 (5.44) & \bf 0.880 (1.78) & \bf 0.947 (1.96)  & 1.400 (1.67) \\
\bottomrule
\end{tabular}
}
\end{table*}

\subsection{Representation Qualities} \label{sec:repr_qual}

To assess the quality of the learned representation, we compute an instance-level representation of each time series by max-pooling representations $\mathbf{r}_t^{(i)}$ along the time so that it can capture general characteristics of the given time series.
We visualize the representations on a two-dimensional plane using t-SNE\cite{van2008visualizing}, one of the most prominent non-linear dimensionality reduction methods.

As illustrated in Figure~\ref{fig:tsne}, on both \texttt{E-Commerce} and \texttt{M5} datasets, our representation module places instances with similar time series features — such as average daily sales ($\equiv \text{MEAN}_t (y_t)$) or coefficient of variation ($\equiv \text{STD}_t (y_t) / \text{MEAN}_t (y_t)$, CoV) — closer together.
As such, our framework is capable of learning useful representations to distinguish different time series even without using any self-supervised learning technique.

\subsection{Forecasting Accuracy and Robustness} \label{sec:acc}

We compare Forchestra to prominent single model based baselines: Auto ARIMA \cite{pmdarima}, Prophet (Meta) \cite{taylor2018forecasting}, Simple Moving Average (SMA), DeepAR (Amazon) \cite{salinas2020deepar}, MQCNN (Amazon) \cite{wen2017multi}, MLP, LSTM \cite{hochreiter1997long}, and Transformer \cite{vaswani2017attention}.
We further test widely used ensemble methods, including Top-K averaging methods ($K = \{ 1, 2, 5, 10, \text{all} \}$) and weighted ensemble methods \cite{gastinger2021study, pawlikowski2020weighted}, by using baselines mentioned above as base models.

Due to the lack of space, the ensemble method with the best test score is reported.
In addition, we implement the simplest version of Deep Ensembles \cite{fort2019deep} by averaging predictions from LSTMs (with the same structure as the base predictor of Forchestra) that is trained with a different random initialization.
We also compare our model to N-BEATS \cite{oreshkin2019n} with ensemble size $100$.
We evaluate the models using a scale-free error metric, mean absolute scale error (MASE) \cite{hyndman2006another}, while providing mean absolute error (MAE) and root mean square error (RMSE) of predictions as supplements.
See Appendix~\ref{sec:hyper} and \ref{sec:eval} for details of model configurations and the evaluation method, respectively. 

Primarily, as illustrated in Table~\ref{tab:comp_train}, our model outperforms competing models by a significant margin.
Furthermore, Forchestra has a relatively low standard deviation across all metrics compared to single model based methods, a desirable property classically attributed to ensembling methods, implying that Forchestra is not only accurate but also reliable in forecasting a wide range of products which is a highly desirable attribute in real-world business applications.
We also observe that a model solely based on self-supervised representations (i.e., TS2Vec) fails to achieve low MASE and MAE but shows the overall lowest standard deviation, demonstrating a high level of robustness.
Based on these findings, we believe Forchestra is capable of making accurate predictions by learning the base predictor more effectively than previous ensemble methods, while also demonstrating the robustness that ensemble methods and representation learning show.
\subsection{Transferability} \label{sec:transfer}
We further assess the transferability of the proposed method on \texttt{E-Commerce} (5.3K) and \texttt{M5} (1.5K) hold-out datasets (i.e., on the unseen test period of unseen products).
We compared our method to supervised models that are trained on the history of the downstream dataset and pre-trained models that are trained on the original dataset used in Section~\ref{sec:acc}.
The results are reported in Table~\ref{tab:comp_test}.
Despite the fact that Forchestra did not observe any time series from the downstream dataset during its training, it maintains a level of accuracy comparable to the one from the training dataset (MASE : $0.880 \rightarrow 0.882$ on \texttt{E-Commerce}) and outperforms supervised models that were trained on the downstream dataset (MASE: $0.882$ vs. $0.912$ of the best among supervised models on \texttt{E-Commerce}) significantly.
It is also worth noting that Forchestra does not suffer from overfitting and is successfully generalizing to unseen datasets despite having 0.8 billion parameters.

\subsection{Scalability}
One of the core research questions we attempt to explore in this paper is the scalability of the proposed framework.
To answer this question, we evaluate Forchestra with different numbers of base predictors, $K = \{ 2, 5, 10, 50, 100 \}$, on the \texttt{E-Commerce} and \texttt{M5} datasets.
As illustrated in Appendix~\ref{sec:hyper}, we use $4$-layer LSTMs with $\textit{hidden\_size}=512$ for base predictors.
In addition, we make a comparison with Forchestra to $400$-layer LSTM and $4$-layer LSTM with $\textit{hidden\_size}=5120$, which have nearly the same number of parameters.

\begin{figure}[t]
    \centering
    \includegraphics[width=0.9\linewidth]{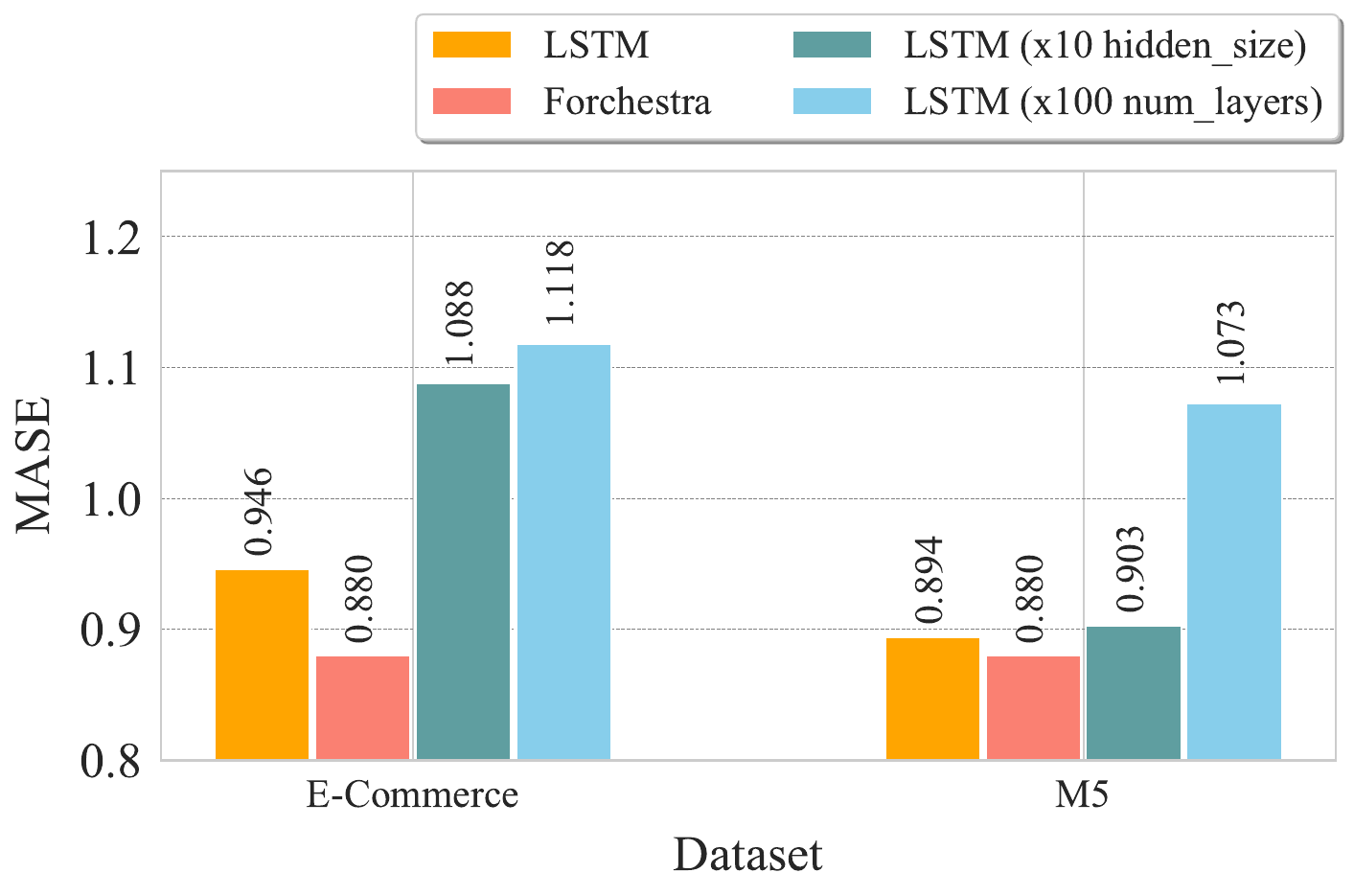}
    \caption{MASE of 4-layer LSTM and different 0.8B models on \texttt{E-Commerce} and \texttt{M5} datasets. Only Forchestra is scalable.}
    \label{fig:scalability}
\end{figure}

\begin{figure}[t]
    \centering
    \subfigure[]
    {
        \includegraphics[width=0.65\columnwidth]{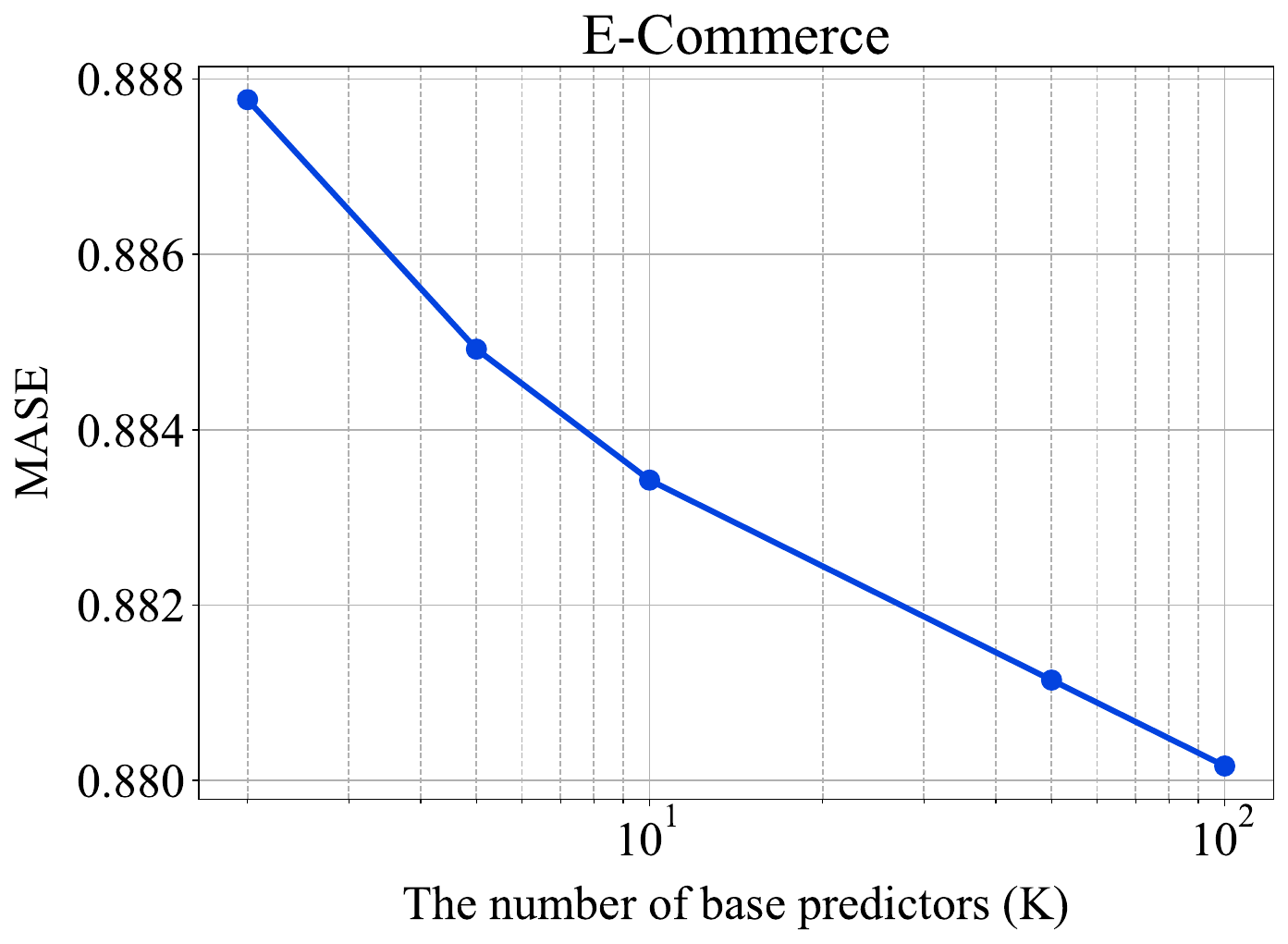}
    }
    \subfigure[]
    {
        \includegraphics[width=0.65\columnwidth]{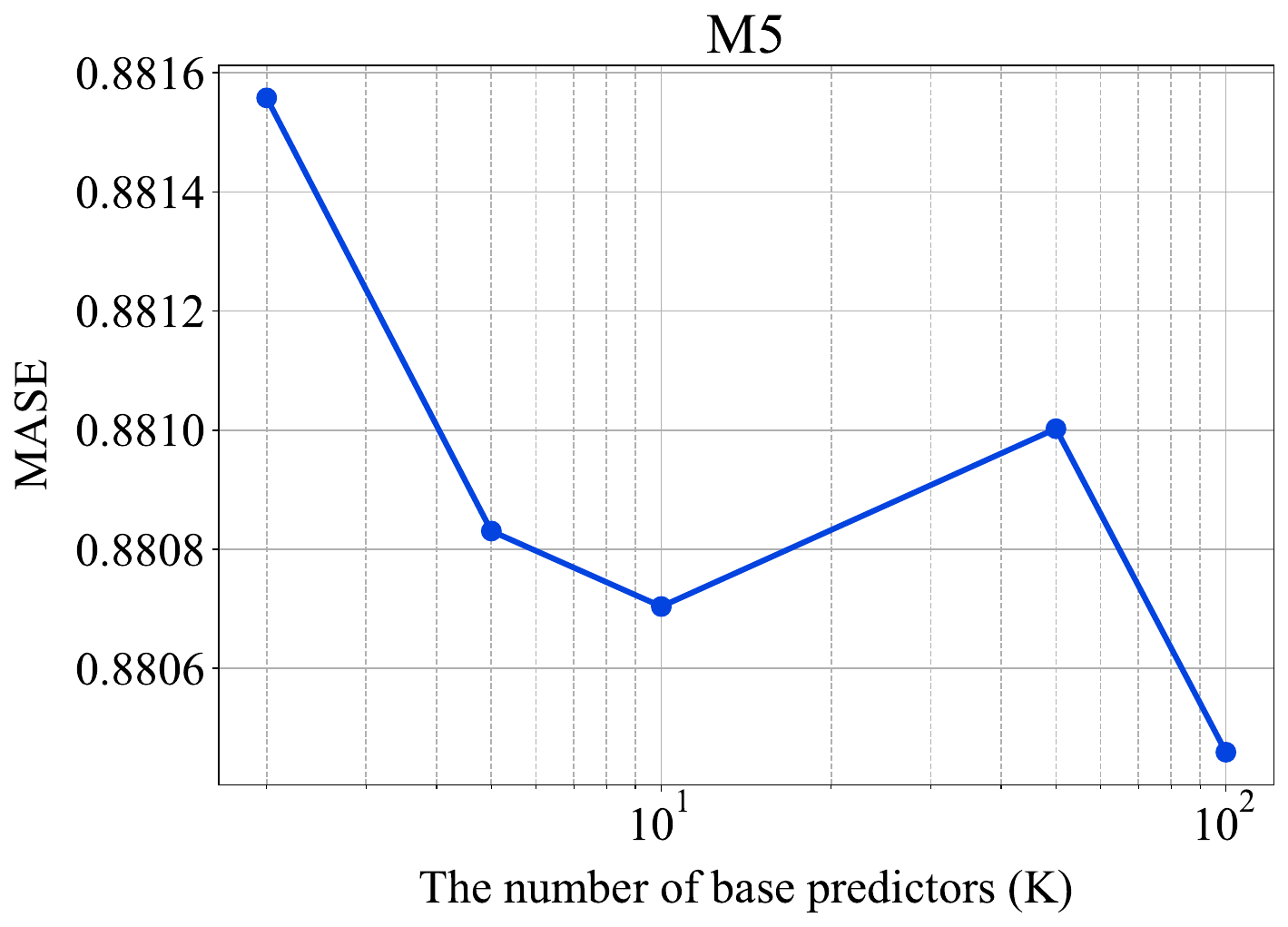}
    }
    \caption{MASE by the number of base predictors (K) on (a) \texttt{E-Commerce} and (b) \texttt{M5} datasets.}
    \vspace{-1.5em}
    \label{fig:scaling_law}
\end{figure}

As shown in Figure~\ref{fig:scaling_law}, we observe that the proposed framework's performance tends to improve as $K$ increases, and Forchestra performs best on both datasets with $K=100$.
The number of learnable parameters of Forchestra with $K=100$ is $842,301,076$ (0.8 billion).
In contrast, as illustrated in Figure~\ref{fig:scalability}, simply increasing the parameter size of the base predictors degraded the performance.

\begin{table*}[t!]
\centering
\caption{Forecast accuracy on downstream dataset (i.e., unseen products). \\ Models performing zero-shot inference are trained on the original dataset.}
\label{tab:comp_test}
{\small
\begin{tabular}{r@{\,}ll|ccc|ccc}
\cmidrule[1.0pt]{2-9}
\multirow{2}{*}{}         & & \multirow{2}{*}{Model} & \multicolumn{3}{c|}{\texttt{E-Commerce} (Unseen Products)} & \multicolumn{3}{c}{\texttt{M5} (Unseen Products)}  \\
                          & &                        & MASE & MAE & RMSE         & MASE & MAE & RMSE       \\
\cmidrule{2-9}
\ldelim\{{12}{*}[\color{black} {\rotatebox[origin=c]{90}{trained on downstream dataset}}\hspace{.45em}]\hspace{.3em}
 & \multirow{3}{*}{Local} & ARIMA                  & 1.045 (1.84) & 0.987 (3.83)  & 1.202 (4.01) & 1.108 (1.67) & 1.163 (2.36)  & 1.482 (2.17) \\
                          & & Prophet                & 1.112 (1.97) & 1.126 (4.21)  & 1.356 (4.34) & 1.125 (1.66) & 1.172 (2.08)  & 1.501 (1.86) \\
                          & & SMA         & 1.145 (2.15) & 1.179 (4.60)  & 1.396 (4.69) & 1.180 (1.92) & 1.413 (2.75)  & 1.843 (2.48) \\
\cmidrule{2-9}
& \multirow{6}{*}{Global} & DeepAR             & 0.932 (2.21) & 0.935 (4.04)  & 1.248 (4.18) & 0.992 (1.94) & 1.147 (2.45)  & 1.643 (2.15) \\
                          & & MQCNN                  & 0.912 (2.07) & 0.916 (3.99)  & 1.222 (4.12) & 0.966 (1.90) & 1.113 (2.41)  & 1.553 (2.16) \\
                          & & MLP                    & 0.945 (2.23) & 1.025 (4.58)  & 1.356 (4.65) & 0.953 (1.87) & 1.069 (2.15)  & 1.521 (1.85) \\
                          & & LSTM                   & 0.920 (2.11) & 0.953 (4.16)  & 1.270 (4.26) & 0.948 (1.87) & 1.065 (2.18)  & 1.510 (1.90) \\
                          & & Transformer            & 1.001 (2.23) & 1.110 (4.47)  & 1.460 (4.54) & 0.952 (1.88) & 1.072 (2.20)  & 1.526 (1.91) \\
                          & & TS2Vec                 & 1.106 (1.86) & 1.101 (4.61)  & 1.302 (4.70) & 1.088 (1.60) & 1.091 (1.84)  & \bf 1.392 (1.63)   \\
\cmidrule{2-9}
& \multirow{3}{*}{\makecell[l]{Ensemble}} & Ensemble (Best)   & 0.908 (2.07) & 0.929 (4.14)  & 1.246 (4.25) & 0.969 (1.72) & 1.021 (1.91)  & 1.399 (1.66) \\
                          & & Deep Ensembles   & 0.930 (2.09) & 0.957 (4.19)  & 1.279 (4.30) & 1.101 (1.76) & 1.152 (2.24)  & 1.580 (1.96) \\
                          & & N-BEATS & 0.889 (2.03) & 0.879 (3.89)  & 1.197 (4.03) &  0.945 (1.90) & 1.062 (2.24)  & 1.525 (1.96) \\
\cmidrule{2-9}  \morecmidrules
\cmidrule{2-9} \ldelim\{{10}{*}[\color{black} {\rotatebox[origin=c]{90}{ zero-shot inference}}\hspace{.45em}]\hspace{.3em} & \multirow{6}{*}{Global} & DeepAR                 & 0.931 (2.23) & 0.943 (4.13)  & 1.267 (4.25) & 0.990 (1.95) & 1.139 (2.37)  & 1.636 (2.06) \\
                          & & MQCNN                    & 0.913 (2.08) & 0.914 (4.03)  & 1.214 (4.17) & 0.957 (1.86) & 1.077 (2.23)  & 1.511 (1.96) \\
                          & & MLP                    & 0.910 (2.03) & 0.943 (4.14)  & 1.260 (4.24) & 0.954 (1.90) & 1.069 (2.21)  & 1.537 (1.91) \\
                          & & LSTM                   & 0.921 (2.02) & 0.950 (4.21)  & 1.245 (4.32) & 0.946 (1.92) & 1.063 (2.22)  & 1.537 (1.93) \\
                          & & Transformer            & 1.021 (2.14) & 1.117 (4.45)  & 1.422 (4.53) & 0.958 (1.93) & 1.079 (2.20)  & 1.559 (1.89) \\
                          & & TS2Vec                 & 1.116 (1.89) & 1.110 (4.48)  & 1.321 (4.57) & 1.082 (1.60) & 1.090 (1.86)  & 1.394 (1.65) \\
\cmidrule{2-9}
& \multirow{3}{*}{\makecell[l]{Ensemble}} & Ensemble (Best)   & 0.900 (2.02) & 0.904 (3.97)  & 1.210 (4.11) & 0.906 (1.86) & 0.983 (1.96)  & 1.444 (1.66) \\
                          & & Deep Ensembles   & 0.946 (1.95) & 0.928 (4.02)  & 1.208 (4.16) & 0.945 (1.89) & 1.061 (2.20)  & 1.511 (1.92) \\
                          & & N-BEATS & 0.892 (2.02) & 0.886 (3.89)  & 1.197 (4.04) & 0.946 (1.90) & 1.063 (2.23)  & 1.521 (1.94) \\
\cmidrule{2-9}
\cmidrule{2-9}
& \bf Ours                      & \bf Forchestra    & \bf 0.882 (1.96) & \bf 0.875 (3.85)  & \bf 1.183 (4.00) & \bf 0.893 (1.84) & \bf 0.965 (1.89)  & 1.408 (1.59) \\
\cmidrule[1.0pt]{2-9}
\end{tabular}
}
\end{table*}

\subsection{Effectiveness of Initialization}\label{sec:effectinit}
\begin{table*}[ht]
\centering
\caption{Ablation studies on Forchestra: Effectiveness of initialization.}
\label{tab:ablation}
{\small
\begin{tabular}{l|ccc|ccc}
\toprule
\multirow{2}{*}{Model} & \multicolumn{3}{c|}{\texttt{E-Commerce}} & \multicolumn{3}{c}{\texttt{M5}}  \\
                      & MASE & MAE & RMSE         & MASE & MAE & RMSE       \\
\midrule
\bf Forchestra &  \bf 0.880 (2.87) & \bf 0.920 (5.69)  & \bf 1.233 (5.44) & \bf 0.880 (1.78) & \bf 0.947 (1.96)  & \bf 1.400 (1.67) \\
\midrule
Freezing rep. module of NC & 0.882 (2.88) & 0.922 (5.72)  & 1.238 (5.46) & 0.882 (1.78) & 0.949 (1.97)  & 1.405 (1.68) \\
\makecell[l]{Freezing rep. module of NC \\ + w/o Pre-training BPs}       & 0.882 (2.88) & 0.927 (5.72)  & 1.244 (5.47) & 0.886 (1.78) & 0.957 (2.00)  & 1.411 (1.71) \\
\midrule
\makecell[l]{w/o Pre-training BPs \& NC \\ ($\sim$ Mixture-of-Experts)}  & 0.882 (2.89) & 0.925 (5.80)  & 1.246 (5.54) & 0.889 (1.79) & 0.960 (2.01)  & 1.426 (1.71) \\
w/o Pre-training NC         & 0.883 (2.88) & 0.924 (5.71)  & 1.238 (5.46) & 0.886 (1.78) & 0.957 (1.99)  & 1.415 (1.70) \\
w/o Pre-training BPs        & 0.884 (2.90) & 0.930 (5.82)  & 1.256 (5.56) & 0.895 (1.79) & 0.973 (2.03)  & 1.435 (1.74) \\
\bottomrule
\end{tabular}
}
\end{table*}

In order to demonstrate the effectiveness of the initialization for the base predictors and the neural conductor, we conduct multiple ablation studies on Forchestra.
The result is shown in Table~\ref{tab:ablation}.

First of all, we freeze the representation module of the neural conductor and train the base predictors, $\{f_{\theta_k}\}_{k=1}^{K}$ and the meta learner of the neural conductor $h_{\phi_w}$.
It is worth noting that the resulting performance is comparable to the original Forchestra, implying that the pre-trained self-supervised representation module is fairly capable of providing expressive and meaningful information to classify the time series appropriately.
Likewise, we believe that the TS2Vec initialization for Forchestra stabilizes the training and reinforces the final performance. 

In addition, we evaluate the performance without pre-training for base predictors and neural conductors (i.e., mixture-of-experts) and report the results in the bottom rows of Table ~\ref{tab:ablation}.
The pre-training phase appears to be beneficial in achieving the best possible performance for Forchestra.
Here we observe a puzzling result; training from randomly initialized BPs but pre-trained NC performs worse than training from both randomly initialized BPs and NC. 
Given that it performs better when the NC is frozen, we suspect that randomly initialized BPs disturb the NC and prevent the model from converging to a better optimum.
We leave this for future research and we think that better initialization methods for the BPs and the NC are interesting topics to explore.
Nonetheless, the models which receive no or only partial initialization are superior compared to other competing baselines.

\section{Discussion} \label{sec:discussion}
\subsection{How are the base predictors constructed?}
\begin{figure*}[t!]
    \centering
    \subfigure[]
    {
        \includegraphics[width=0.65\columnwidth]{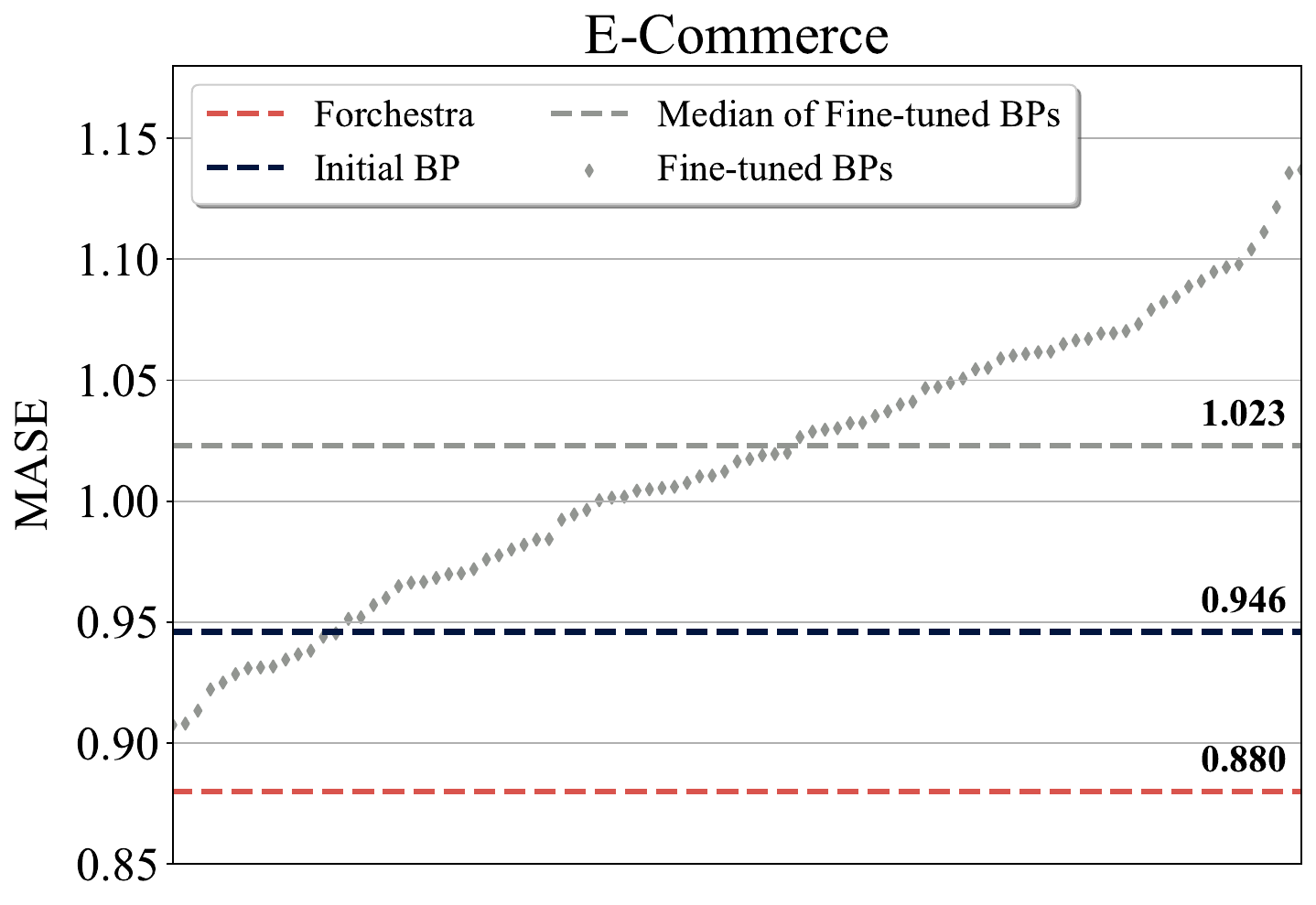}
    }
    \subfigure[]
    {
        \includegraphics[width=0.65\columnwidth]{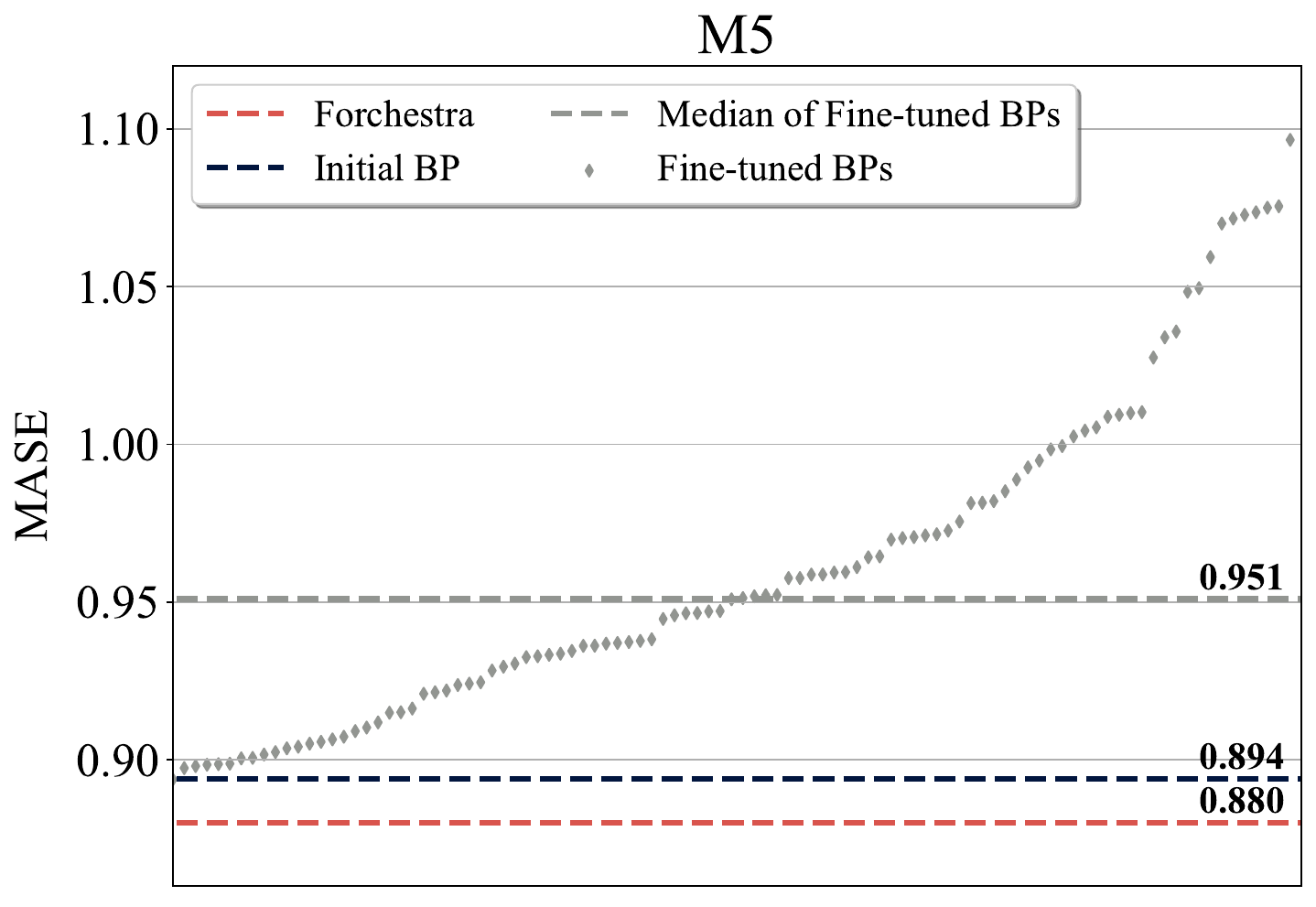}
    }
    \caption{MASE of fine-tuned base predictors on (a) \texttt{E-Commerce} and (b) \texttt{M5} datasets.}
  \label{fig:bp_analysis}
\end{figure*}

Existing ensemble approaches gain benefits by forming a pool of strong base models. 
To see if Forchestra operates with such a rationale, we evaluate the fine-tuned base predictors and report the average MASE on the original dataset in Figure~\ref{fig:bp_analysis}.
The performance of the BP used for initialization, which is equivalent to a 4-layer LSTM, is reported as reference.
Noteworthily, the performance level of the fine-tuned BPs is far from the final score achieved by Forchestra. 
They perform on average even worse than the initial BP. 
This finding shows that Forchestra does not simply construct a set of strong base models.
In the following section, we investigate how Forchestra can achieve superior performance by aggregating such weak predictors.

\begin{table*}[ht]
\centering
\caption{Comparison of forecast accuracy between Forchestra and ensemble approaches. \\ The highest scores among ensemble methods are underlined.}
\label{tab:ens}
{\small
\begin{tabular}{ll|ccc|ccc}
\toprule
\multirow{2}{*}{Approach}         & \multirow{2}{*}{Model} & \multicolumn{3}{c|}{\texttt{E-Commerce}} & \multicolumn{3}{c}{\texttt{M5}}  \\
                          &                        & MASE & MAE & RMSE         & MASE & MAE & RMSE       \\
\midrule
\bf Orchestra                  & \bf Forchestra &  \bf 0.880 (2.87) & \bf 0.920 (5.69)  & \bf 1.233 (5.44) & \bf 0.880 (1.78) & \bf 0.947 (1.96)  & \bf 1.400 (1.67) \\
\midrule
\multirow{3}{*}{\makecell[l]{Divide-and- \\ Conquer}} 
  &  DnC & 0.899 (2.93) & 0.953 (5.94)  & 1.274 (5.67) & 0.889 (1.78) & 0.962 (2.00)  & 1.420 (1.71) \\
  & {Forch-DnC} & 0.887 (2.88) & 0.938 (5.75)  & 1.245 (5.50) & 0.886 (1.78) & 0.955 (1.99)  & 1.409 (1.70) \\
  & {Forch-DnC+} &  0.880 (2.88) & 0.920 (5.71)  & 1.236 (5.46) & 0.881 (1.78) & 0.947 (1.97)  & 1.403 (1.67) \\
\midrule
\multirow{9}{*}{\makecell[l]{Ensemble \\ (global)}} 
                          & Average    & 0.905 (2.92) & 0.980 (5.96)  & 1.313 (5.69) & 0.890 (1.79) & 0.962 (2.03)  & 1.432 (1.73)) \\
                          & Top-50 & 0.922 (2.95) & 0.995 (5.98)  & 1.333 (5.70) & \underline{0.886 (1.79)} & \underline{0.955 (1.99)}  & 1.415 (1.70) \\
                          & Top-10 & 0.906 (2.92) & 0.982 (6.00)  & 1.311 (5.73) & 0.889 (1.79) & 0.959 (1.98)  & \underline{1.413 (1.68)} \\
                          & Top-5 & 0.905 (2.92) & 0.974 (5.92)  & 1.303 (5.65) &  0.890 (1.80) & 0.961 (2.00)  & 1.417 (1.70) \\
                          & Top-2 & \underline{0.904 (2.92)} & \underline{0.964 (5.82)}  & \underline{1.293 (5.56)} & 0.893 (1.79) & 0.965 (1.99)  & 1.414 (1.70) \\
                          & Top-1 & 0.908 (2.92) & 1.004 (6.08)  & 1.334 (5.81) & 0.894 (1.79) & 0.971 (2.04)  & 1.429 (1.75) \\
                          & W-inv & 0.914 (2.93) & 0.992 (5.98)  & 1.327 (5.71) & 0.890 (1.79) & 0.962 (2.03)  & 1.431 (1.73) \\
                          & W-sqr & 0.920 (2.94) & 1.000 (5.99)  & 1.337 (5.72) & 0.889 (1.79) & 0.961 (2.02)  & 1.430 (1.72) \\
                          & W-exp & 0.914 (2.93) & 0.991 (5.98)  & 1.327 (5.71) & 0.890 (1.79) & 0.962 (2.03)  & 1.431 (1.73) \\
\midrule
\multirow{8}{*}{\makecell[l]{Ensemble \\ (instance-wise)}} 
                          & Top-50 & 0.916 (2.93) & 0.977 (5.89)  & 1.318 (5.62) & 0.894 (1.79) & 0.964 (1.99)  & 1.421 (1.70) \\
                          & Top-10 & 0.916 (2.93) & 0.973 (5.82)  & 1.302 (5.55) & 0.902 (1.78) & 0.980 (2.02)  & 1.436 (1.73) \\
                          & Top-5 & 0.924 (2.95) & 0.986 (5.87)  & 1.314 (5.61) & 0.910 (1.79) & 0.993 (2.04)  & 1.449 (1.75) \\
                          & Top-2 & 0.944 (2.98) & 1.014 (5.95)  & 1.344 (5.68) & 0.929 (1.81) & 1.019 (2.09)  & 1.480 (1.79) \\
                          & Top-1 & 0.969 (3.05) & 1.057 (6.16)  & 1.390 (5.88) & 0.949 (1.84) & 1.049 (2.16)  & 1.521 (1.86) \\
                          & W-inv & 0.937 (3.47) & 1.120 (22.25)  & 1.449 (21.32) & 0.889 (1.79) & 0.959 (2.01)  & 1.426 (1.71) \\
                          & W-sqr & 0.939 (3.48) & 1.118 (22.24)  & 1.449 (21.31) & 0.889 (1.79) & 0.958 (2.01)  & 1.424 (1.71) \\
                          & W-exp & 0.932 (3.47) & 1.116 (22.25)  & 1.444 (21.32) & 0.889 (1.79) & 0.959 (2.01)  & 1.427 (1.72) \\
\bottomrule
\end{tabular}
}
\end{table*}

\begin{table*}[ht]
\centering
\caption{Rank-biased overlap (RBO) score between Forchestra and ensemble approaches.
}
\label{tab:ranking}
{\small
\begin{tabular}{l|ccc|ccc}
\toprule
\multirow{2}{*}{Model} & \multicolumn{3}{c|}{\texttt{E-Commerce}} & \multicolumn{3}{c}{\texttt{M5}}  \\
                       & RBO@5 & RBO@10 & RBO@50 & RBO@5 & RBO@10 & RBO@50       \\
\midrule
Forchestra & 0.0093 & 0.0294 & 0.0931 & 0.0183 & 0.0390 & 0.1058 \\
\midrule
\makecell[l]{Ranked by val. score (global)} & 0.0148 & 0.0363 & 0.1013 & 0.0112 & 0.0352 & 0.1121 \\
\midrule
\makecell[l]{Ranked by val. score (instance-wise)} & 0.1770 & 0.3085 & 0.5317 &  0.0511 & 0.1030 & 0.2249 \\
\bottomrule
\end{tabular}
}
\end{table*}

\subsection{How does the conductor guide predictors?}
What makes the proposed orchestra approach different from classical ensemble approaches that boost performance?
Does the neural conductor classify time series with the representation module and appropriately delegate tasks to the base predictors (H1)?
Or is the neural conductor of Forchestra simply better at selecting the base predictor that is likely to perform well in the future (H2)?

To explore H1, we implement the divide-and-conquer (DnC) method as follows.
We first use K-means clustering on the self-supervised representation by the pre-trained TS2Vec to divide the time series instances into $K=100$ groups.
Then, for each cluster, we assign a base predictor (which is initialized using the same method as in Forchestra) so that each predictor can be fine-tuned and predict the future (i.e., each one acts as an expert in their respective cluster).
In addition, by using the same $K=100$ base predictors trained from the Forchestra framework, we compare our model with classical ensemble approaches including Top-K and weighted ensemble methods (see Section~\ref{sec:ensemble} for details).
If H1 is correct, then a divide-and-conquer strategy should be able to match Forchestra's performance and perform better than other ensemble methods.

As shown in Table~\ref{tab:ens}, however, we discover that neither a divide-and-conquer strategy nor ensemble methods were successful in forecasting as accurate as Forchestra.
Despite the fact that the DnC method outperforms other baselines, no clear superiority or inferiority between DnC and the best ensemble method could be demonstrated.
We also report the scores of two different Forchestras, in which the base predictors are initialized to those of DnC and then frozen (Forch-DnC) or fine-tuned (Forch-DnC+); but performance is nearly identical or worse than the original Forchestra.
This implies once again that the divide-and-conquer strategy is not how Forchestra operates.
Hence, we reject H1.

To investigate H2, we design an experiment as follows.
First, we evaluate each base predictor for each time series instance on the test period to find the ground truth rank among them.
Similarly, we instance-wisely (i.e., locally) and globally rank base predictors on validation period as conventional ensemble methods do.
Then we analyze the weight vectors $\mathbf{w}_t$ in Equation~\ref{eq:meta} inferred by the neural conductor of Forchestra and calculate the estimated rank by comparing the assigned weight value of each base predictor.
Finally, we measure the similarity among those ranks by rank-biased overlap (RBO) \cite{webber2010similarity}.
The higher the RBO is, the more similar the two rankings are.
If H2 is correct, Forchestra should show a higher RBO than conventional ensembling methods, implying that Forchestra is able to make a choice among the BPs that performs better during the test period than simply taking the best ones from the validation period.
Surprisingly, the RBO score of Forchestra is the lowest, as shown in Table~\ref{tab:ranking}.
Furthermore, instance-wise ensemble methods have the highest RBO, but they perform worse than Forchestra or global ensemble methods.
As such, identifying the best predictors does not guarantee the best results.
Therefore, we reject H2 as well, concluding with: "Unity is strength; Forchestra learns to harmonize."

\section{Conclusion and Future Work}
\label{sec:futurework}

In this paper, a simple but powerful time series prediction framework for demand forecasting was presented.
The proposed framework consists of a number of base predictors operated by the neural conductor that adaptively adjusts the importance weights of base predictors based on the target item's representation.
Experiments showed that the proposed method not only outperformed single-model approaches and prominent ensemble methods, but also was transferable to smaller downstream datasets of unseen products in a zero-shot fashion.
The model size was scalable up to $K=100$ predictors (0.8 billion parameters).
The paper also investigated the difference between conventional ensemble approaches and the proposed framework in terms of base model selection strategy.

In future research, larger scale frameworks with thousands of base predictors can be further explored. The authors also think that an enhancement to the representation module could turn out to be an interesting avenue for future work: One idea is to add an auxiliary loss (e.g., self-supervised contrastive loss) to the representation module. Another idea is based on the observation that the currently trained representations are unable to cluster categories (see Figure ~\ref{fig:tsne}), which classically are thought to be important features in time series forecasting. Therefore approaches that build a representation on top of multimodal data (time series, categories, product name text embeddings, etc.) could potentially result in an even more capable neural conductor.

\bibliographystyle{IEEEtran}
\bibliography{bigforecast_bibliography}

\clearpage
\appendices
\section{Experiment Details} \label{sec:hyper}

All deep learning models take an input of $context\_length (C) = 28 \times 3 = 84$.
For models with multiple hyperparameters, the one with best validation score is selected and evaluated on the test dataset.

\subsection{Classical Models}
\begin{itemize}[leftmargin=12pt]
    \item Auto ARIMA \cite{pmdarima}: $m=7$ is applied following the recommendations from the reference. For others, we followed the package's default options.
    \item Prophet \cite{taylor2018forecasting}: We followed the package's default options.
    \item Simple Moving Average: We tested with \textit{look\_back\_period} $ = \{ 7, 14, \ldots, 70 \}$.
\end{itemize}

\subsection{Deep Learning Models} \label{sec:model}
The paper trained and evaluated deep learning models implemented by GluonTS \cite{alexandrov2019gluonts} or PyTorch \cite{NEURIPS2019_9015} as baselines.
\begin{itemize}[leftmargin=12pt]
    \item DeepAR \cite{salinas2020deepar}: In the same manner as in Forchestra, \textit{cell\_type} $ = lstm$, and \textit{num\_cells} $=512$. We tested with \textit{num\_layers} $ = \{ 1, 4, 16 \}$ and \textit{distr\_output} = \{StudentTOutput(), NegativeBinomialOutput()\}. For others, we followed the GluonTS's default options. 
    \item MQCNN \cite{wen2017multi}: We followed the GluonTS's default options.
    \item N-BEATS \cite{oreshkin2019n}: We evaluated N-BEATS with \textit{meta\_bagging\_size} $=10$, \textit{meta\_context\_length} $ = \{ 21, 28, \ldots, 84 \}$, \textit{meta\_loss\_function} = MASE so that the total number of models for its ensemble is $10 \times 10 \times 1 = 100$ as in Forchestra. For others, we followed the GluonTS's default options.
    \item MLP: The first layer of MLP model takes a $\mathbf{x}_{t-C-1:t}$ as input and outputs $512$ dimensional hidden states. The hidden states then pass through additional \textit{num\_layers} $ = \{ 1, 4, 16 \}$ layers with ReLU activations. The final output size is $P$. %
    \item LSTM \cite{hochreiter1997long}: We evaluated LSTM with \textit{hidden\_size} $=512$ and \textit{num\_layers} $ = \{ 1, 4, 16 \}$. The output features from the last layer of the LSTM are then transformed into $P$ dimensional output by a fully-connected layer.
    \item Transformer \cite{vaswani2017attention}: We evaluated Transformer with \textit{d\_model} $=512$, \textit{dim\_feedforward} $ = 512 \times 4$, \textit{nhead} $=4$ and \textit{num\_layers} $ = \{ 1, 4, 16 \}$. The output features from the last layer of the model is then transformed into $P$ dimensional output by a fully-connected layer.
    \item TS2Vec \cite{yue2021ts2vec}: The first linear projection layer takes $\mathbf{x}_{t-W-1:t}$ as input and maps each timestamp $\mathbf{x}_{i}$ to a high-dimensional latent vector $\mathbf{z}_{i}$ with \textit{projection\_dim} $=64$, where $W=200$. Then dilated CNN modules with \textit{num\_residual\_blocks} $=5$, \textit{kernel\_size} $=3$, \textit{output\_dims} $=32$ takes a latent vector and extract a representation $\mathbf{r}_{i}$ at each timestamp. The additional linear regression model takes last timestamp representation $\mathbf{r}_{i,t}$  and transform it into $P$ dimensional output. We used the official implementation~\footnote{\url{https://github.com/yuezhihan/ts2vec}}. 
\end{itemize}

\subsection{Forchestra}
\begin{itemize}[leftmargin=12pt]
    \item Base Predictors: We used $4$-layer LSTMs with $\textit{hidden\_size}=512$ of Section~\ref{sec:model} as the base predictors. We tested with $K=\{2, 5, 10, 50, 100\}$.
    \item Neural Conductor: We used the same dilated CNN of TS2Vec for the representation module and a fully connected layer for the meta learner.
\end{itemize}

\subsection{Ensemble Methods} \label{sec:ensemble}

\begin{itemize}[leftmargin=12pt]
    \item Top-K: The ensemble approach takes the $K$ best performing models based on the validation score (i.e., MASE). Then it averages $K$ predictions from those for the test period. We tested with $K=\{1, 2, 5, 10, \text{min}(50, \text{all})\}$.
    \item Weighted Ensemble \cite{gastinger2021study}: The ensemble approach evaluates the validation score ($S$) for all base models. Then it averages $K$ predictions with weights calculated by following functions introduced in \cite{pawlikowski2020weighted}:
    \begin{enumerate}[leftmargin=30pt]
        \item \textbf{W-inv}: ~ $g_{inv}(S) = 1 / (S + \epsilon)$
        \item \textbf{W-sqr}: ~ $g_{sqr}(S) = g_{inv}(S)^2$
        \item \textbf{W-exp}: ~ $g_{exp}(S) = \exp(g_{inv}(S))$
    \end{enumerate}
    where a small epsilon ($= 1\mathrm{e}{-10}$) is added to the denominator to avoid division by zero.
    \item Deep Ensembles \cite{fort2019deep}: We trained same $100$ LSTMs as of Forchestra. Each model is randomly initialized. After training, it averages all predictions for the test period.
\end{itemize}

For Top-K and weighted ensemble methods, we applied ensemble both globally and locally (i.e., instance-wisely).\\
\noindent\textbf{Global method}: scores each base predictor based on its average performance across all time-series instances, and aggregate them with the same weight for each instance.\\
\noindent\textbf{Local method}: scores each predictor based on its performance for each instance.

\section{Evaluation} \label{sec:eval}
We evaluate models based on three metrics: mean absolute scale error (MASE), mean absolute error (MAE), and root mean square error (RMSE).
In real-world demand forecasting platforms, evaluating performance on non-sale days is meaningless, so we revised the evaluation metrics as follows:
\small
\begin{align}
        \text{MAE}(y_t, \hat{y}_t) &= \sum_{t \in \mathcal{T}_p} |y_t - \hat{y}_t| ~ / ~ |\mathcal{T}_p| \\
        \text{RMSE}(y_t, \hat{y}_t) &= \sqrt{\sum_{t \in \mathcal{T}_p} (y_t - \hat{y}_t)^2 ~ / ~ |\mathcal{T}_p|} \\
        \text{MASE}(y_t, \hat{y}_t) &= \frac{\text{MAE}(y_t, \hat{y}_t)}{
                \sum_{t \in \mathcal{T}_h} |y_t - y_{t-1}| ~ / ~ |\mathcal{T}_h|
            }
\end{align}
\begin{gather}
    \mathcal{T}_p = \{t ~ | ~ a_t=\text{sale}, ~ T+1 \le t \le T+P \} \nonumber \\
    \mathcal{T}_h = \{t ~ | ~ a_{t-1:t}=\text{sale}, ~ 2 \le t \le T \} \nonumber
\end{gather}
\normalsize
where $a_t \in \{ \text{sale}, \text{non\_sale} \}$ is availability status at $t$.
Similarly, evaluating products with too few records can lead to a bias in forecasting method performance evaluations as well as zero-division errors in MASE calculations; thus in case of \texttt{E-Commerce}, products with at least 7 days of historical sales (i.e., $|\mathcal{T}_p^{(i)}| \ge 7$) are evaluated for each test period.

\end{document}